# Spatio-temporal-spectral-angular observation model that integrates observations from UAV and mobile mapping vehicle for better urban mapping

Zhenfeng Shao[a], Gui Cheng[a], Deren Li[a], Xiao Huang[b], Zhipeng Lu[c], Jian Liu[c]

[a] State Key Laboratory of Information Engineering in Surveying, Mapping and Remote Sensing, Wuhan University, Wuhan, China; [b] Department of Geosciences, University of Arkansas, Fayetteville, USA; [c] Leador Spatial Information Technology Corporation, Wuhan, China.

ABSTRACT

In a complex urban scene, observation from a single sensor unavoidably leads to voids in observations, failing to describe urban objects in a comprehensive manner. In this paper, we propose a spatio-temporal-spectral-angular observation model to integrate observations from UAV and mobile mapping vehicle platform, realizing a joint, coordinated observation operation from both air and ground. We develop a multi-source remote sensing data acquisition system to effectively acquire multi-angle data of complex urban scenes. Multi-source data fusion solves the missing data problem caused by occlusion and achieves accurate, rapid, and complete collection of holographic spatial and temporal information in complex urban scenes. We carried out an experiment on Baisha Town, Chongqing, China and obtained multi-sensor, multi-angle data from UAV and mobile mapping vehicle. We first extracted the point cloud from UAV and then integrated the UAV and mobile mapping vehicle point cloud. The integrated results combined both the characteristic of UAV and mobile mapping vehicle point cloud, confirming the practicability of the proposed joint data acquisition platform and the effectiveness of spatio-temporal-spectral-angular observation model. Compared with the observation from UAV or mobile mapping vehicle alone, the integrated system provides an effective data acquisition solution towards comprehensive urban monitoring.

CONTACT    Zhenfeng Shao    ✉ shaozhenfeng@whu.edu.cn

# 1. Introduction

With over half the population living in urban agglomerations (Hoole, Hincks, and Rae 2019), urban systems pose unique remote sensing challenges. As a branch of remote sensing technology, urban remote sensing takes the city as the observation object (Weng and Quattrochi 2018, Yang 2011, Shao et al. 2020, Wu, Gui, and Yang 2020, Huang and Wang 2020). Urban fabrics are rather complex, with the existence of various kinds of occlusion from vertical objects such as buildings and trees. Satellite/aerial imagery usually fails to mitigate the lack of urban scene information caused by the occlusions. As a result, the information regarding urban surfaces is difficult to be extracted in an accurate manner, leading to the limited utility of high-resolution remote sensing imagery when it is applied in urban scenes. Therefore, there is a need to carry out both horizontal and vertical spatial observations to form a coordinated observation that involves multiple platforms.

With the rapid development of computational technology and the emergency of the new generation of surveying and mapping technology (e.g., satellite navigation and positioning technology (Ning, Yao, and Zhang 2013), remote sensing and geographic information technology (Dar, Sankar, and Dar 2010, Li, Shao, and Zhang 2020), traditional remote sensing surveying and mapping technology has undergone a fundamental change. With the continuous progress of Unmanned Aerial Vehicle (UAV) (Colomina and Molina 2014) and Mobile Mapping System (MMS) (Petrie 2010), UAV and MMS have established a new venue where spatial information can be retrieved in a timely manner. Together, they play a very important role in the field of remote sensing surveying and mapping. Therefore, we argue that it is of great importance to investigate how to integrate UAV and MMS towards a coordinated urban mapping framework.

UAV, aiming to obtain spatial information within a targeted region, is an emerging remote sensing platform with the capability to carry a variety of remote sensing sensors, such as high-resolution CCD digital camera, light optical camera, multispectral imager, infrared scanner, laser scanner, hyperspectral imager, synthetic aperture radar, to list a few. UAV remote sensing technology (Changchun et al. 2010, Yao, Qin, and Chen 2019, Zhang et al. 2020, Ma et al. 2013, Xia et al. 2018, Shao et al. 2021) has received wide attention, given its several advantages. 1) UAV operations are flexible and efficient, as UAVs do not require large space for taking off and landing. In addition, UAVs are less affected by weather conditions compared to satellite-borne instruments. 2) UAVs are able to obtain multi-scale imagery thanks to their dynamic flight height, allowing them to perform both large-scale and small-scale monitoring. 3) The resolution of UAV images can reach 0.1 meters or even higher, considerably finer than most satellite imagery. 4) The cost of a UAV remote sensing system is much lower than that of satellite remote sensing and aerial remote sensing in terms of platform construction, routine maintenance, and aerial photography. 5) UAV system is highly integrated. A UAV system is generally equipped with a mission planning system and data processing system with simple and flexible flight planning. The data processing can be completed shortly after the data acquisition.

Mobile mapping system (Puente et al. 2013, Li 2006, Novak 1995, Marinelli et al. 2017, Sester 2020), born in the early 1990s, is one of the cutting-edge techniques of modern surveying and mapping. A mobile mapping system is the composition of global satellite positioning, inertial navigation, image processing, photogrammetry, laser scanning, geographic information, and integrated control technology. It has many advantages, such as high flexibility, high precision, high resolution, real-time data acquisition, and multi-source data collection. Mobile mapping system can contribute to urban mapping by obtaining information on surfaces, and more importantly, information on vertical urban objects (e.g., buildings and trees). One specific category of mobile mapping system is the utilization of mobile mapping vehicle. Given its capability of capturing objects' facades, mobile mapping vehicle has been widely used in 3D modeling, topographic map update, GIS database construction, urban survey and planning, mine survey, public security, and urban management.

Although UAV systems and mobile mapping vehicle systems are the current cutting-edge sensing technology, observations based on the single sensor have insurmountable shortcomings in comprehensively describing the urban environments, as single-sensor, either from UAV system or mobile mapping system fails to capture three-dimensional spatial information. For example, UAV can provide the spatial information and texture features of objects from the top view, lacking the details of geometric and texture information of the facades of the objects. The mobile mapping system can obtain street view data with high position accuracy and high resolution, providing rich facade information and a better three-dimensional description of the scene. However, the 3D point cloud data are usually noisy and lacking the texture information from the top view (Haala et al. 2008, Hana et al. 2018). We argue that data collected from multi-platforms can be complementary. In this paper, we focus on the multi-source data fusion methods, taking advantage of multi-sensor acquisition to obtain high-quality geospatial data.

Multi-sensor integration and fusion technology from the early 1980s in the military field have rapidly expanded to military and non-military applications (Huang et al. 2010, Li and Fu 2018). Multi-sensor integration refers to the comprehensive use of information from various sensors obtained in different time spans to assist task completion, including the data collection, transmission, analysis, and synthesis of useful information provided by various sensors. The purpose of multi-sensor integration is to take advantage of the resources from multiple sensors, especially complementary spatiotemporal information, to obtain consistent interpretation and description of measured objects being measured. Several multi-sensor fusion methods have been developed in the remote sensing domain (Li 2017), including spatio-temporal fusion (Huang and Song 2012, Song and Huang 2012, Amorós-López et al. 2013), spatio-spectral fusion (Ma et al. 2020, Xu et al. 2020, Shao, Wu, and Guo 2020, Shao and Cai 2018, Pingxiang and Zhijun 2003), spatio-temporal-spectral fusion (Shao et al. 2019, Shen, Meng, and Zhang 2016).

In a complex urban scene, observation from a single sensor unavoidably results in voids in observations, failing to describe urban objects in a comprehensive manner. In this study, we achieved the rapid acquisition and integration of three-dimensional seamless holographic spatial and temporal data of cities. We integrated observations from UAV and mobile mapping vehicle and proposed a spatio-temporal-spectral-angular observation model (Figure 1). The model we proposed achieves fast observation through the integration of multi-platform, multi-angle observations in both the air and ground. It obtains various data types such as images, point clouds, positions, and attitudes with synchronized and unified geographic reference. Through multi-source data fusion, the spatiotemporal holographic information of complex urban scenes can be collected in rapid, fast, and comprehensive manners.

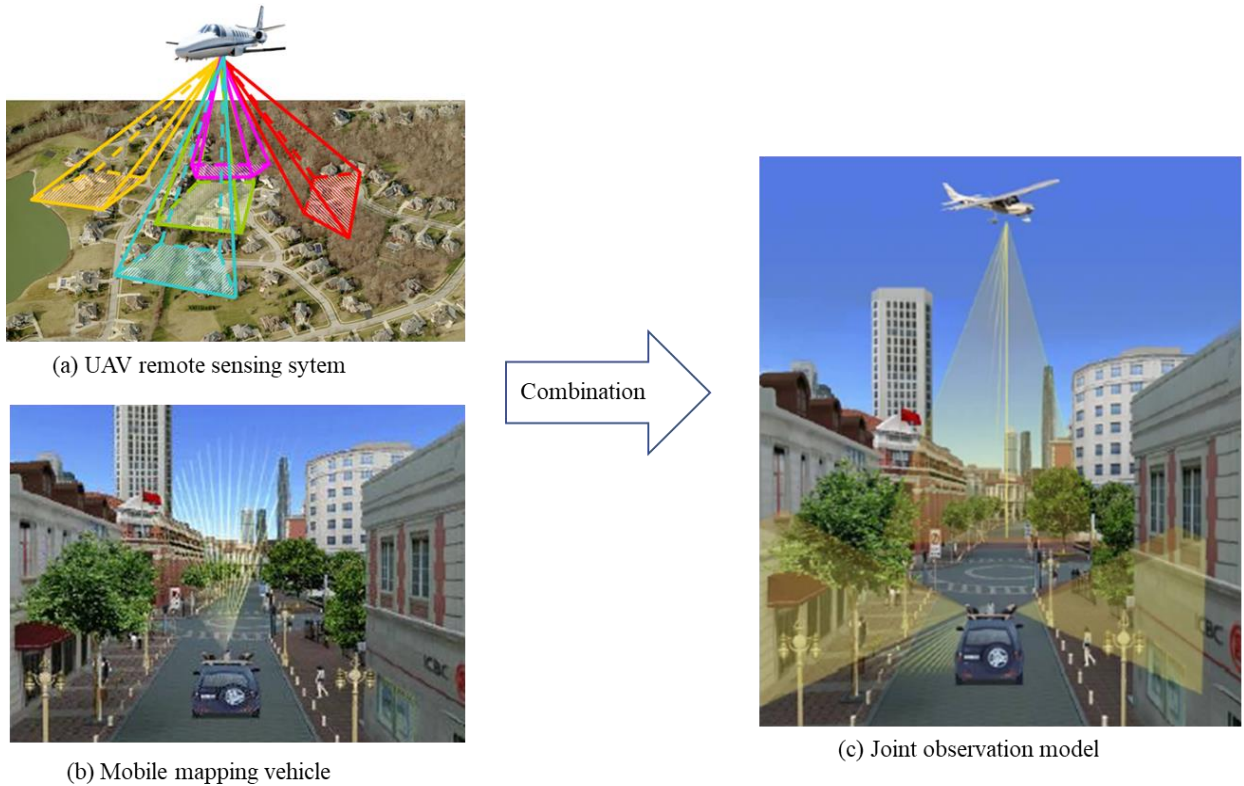

Figure 1. The spatio-temporal-spectral-angular observation model that combines observations from UAV and mobile mapping vehicle.

## 2. Methodology

In this section, we introduce the spatio-temporal-spectral-angular observation model at pixel-level (or voxel-level) and feature-level. Then the 3D point cloud registration is introduced using data collected from UAV and mobile mapping vehicle.

### 2.1 Spatio-temporal-spectral-angular observation model

In this study, we proposed a spatio-temporal-spectral-angular observation model. Theoretically, both the UAV platform and the mobile mapping vehicle platform can be equipped with sensors with high-temporal, high-spatial, and high-spectral resolution. The proposed spatio-temporal-spectral-angular observation model is able to handle the discrepancies in observation angels from these two platforms. Urban remote sensing observation is generally achieved by constructing appropriate models or algorithms based on the temporal, spatial, spectral, and angular image features. Therefore, we abstract the process as a spatio-temporal-spectral-angular model, where temporal, spatial, spectral, and angular features serve as model inputs. According to different model outputs, an urban remote sensing observation model can be generally divided into two categories: 1) data quality improvement model (pixel-level fusion or voxel-level fusion), and 2) information extraction model (feature-level fusion).

1) The data quality improvement model refers to obtaining images with higher quality by fusing multi-source data. This process can be modeled using the following formula:

$$I = O(I_1, I_2, I_3, ..., I_K) \quad (1)$$

where $I_1, I_2, I_3, ..., I_K$ represent multi-source images, $O(\bullet)$ stands for the fusion model, and $I$ indicates the output of the model, i.e., images with improved quality.

In general, remote sensing images mainly contain spatial, spectral, temporal, and angular features, which can be expressed as:

$$I_i = I_{i,spatial} \oplus I_{i,temporal} \oplus I_{i,spectral} \oplus I_{i,angular} \quad (2)$$

where $I_i$ stands for a remote sensing image, $I_{i,spatial}$, $I_{i,temporal}$, $I_{i,spectral}$ and $I_{i,angular}$ represent the sets of spatial, temporal, spectral, and angular features, respectively.

Given the difference in sensing techniques, multi-source images tend to focus only on certain components. For example, high-resolution images own high spatial resolution but usually couples with limited temporal and limited spectral resolution. To fuse these components, the constraint relationship among multi-source images on each component should be established when a spatio-temporal-spectral-angular observation model is constructed:

$$I = O\left(F\left(\{I_{i,spatial}\}_{i=1}^{K}\right) \oplus F\left(\{I_{i,temporal}\}_{i=1}^{K}\right) \oplus F\left(\{I_{i,spectral}\}_{i=1}^{K}\right) \oplus F\left(\{I_{i,angular}\}_{i=1}^{K}\right)\right) \quad (3)$$

where $F(\bullet)$ is the feature constraint function and $I$ is the fused image.

2) When information serves as the output of spatio-temporal-spectral-angular observation models, the information extraction model with a specific task $T$ can be abstracted as:

$$Y = O(I_1, I_2, I_3, ..., I_K; T) \quad (4)$$

where $I_1, I_2, I_3, ..., I_K$ represent multi-source images, $O(\bullet)$ stands for information extraction model, and $Y$ indicates the output of the model.

Similarly, under the constraint of the task $T$, features from four aspects (i.e., spatial, temporal, spectral, and angular)

can be extracted and further fused, thereby outputting useful information that benefits numerous urban monitoring tasks. This process can be expressed via the following formula:

$$Y = O\left(F\left(\{I_{i,spatial}\}_{i=1}^{K};T\right) \oplus F\left(\{I_{i,temporal}\}_{i=1}^{K};T\right) \oplus F\left(\{I_{i,spectral}\}_{i=1}^{K};T\right) \oplus F\left(\{I_{i,angular}\}_{i=1}^{K};T\right)\right) \quad (5)$$

where $F(\bullet)$ is the feature constraint function and $O(\bullet)$ represents the information extraction function.

The spatio-temporal-spectral-angular observation model is a general model for data fusion of similar types. The data quality improvement model performs pixel-level (or voxel-level) fusion, while the information extraction model performs feature-level fusion. In this study, we conduct the voxel-level fusion based on the 3D point cloud data with diversified spatial, temporal, spectral, and angular information.

## 2.2 3D point cloud registration

To perform voxel-level fusion based on the 3D point cloud data from UAV and mobile mapping vehicle, the point cloud registration is a vital preprocess. Point cloud registration refers to the process of transforming the point cloud to the same coordinate system through a series of rotation and translation operations. Point cloud registration belongs to a rigid transformation that the target point $Q$ is transformed to the source point $P$ through rotation and translation transformation that can be represented by Equation. 6, where $R$ and $T$ refer to the rotation matrix and translation matrix respectively, the $M$ refers to the total transformation matrix ($M$) that describes the three-dimensional transformation of space is in the form of 4×4.

$$P = R \bullet Q + T = M \bullet Q \quad (6)$$

Using the three axes of the coordinate system $X$, $Y$, and $Z$ as the axes of rotation, $\theta$, $\alpha$ and $\beta$ are the rotation angle about the axis of $X$, $Y$, and $Z$, respectively. The three rotation matrices are obtained as follows:

$$R_x(\theta) = \begin{bmatrix} 1 & 0 & 0 \\ 0 & \cos\theta & -\sin\theta \\ 0 & \sin\theta & \cos\theta \end{bmatrix} \quad (7)$$

$$R_y(\alpha) = \begin{bmatrix} \cos\alpha & 0 & \sin\alpha \\ 0 & 1 & 0 \\ -\sin\alpha & 0 & \cos\alpha \end{bmatrix} \quad (8)$$

$$R_z(\beta) = \begin{bmatrix} \cos\beta & -\sin\beta & 0 \\ \sin\beta & \cos\beta & 0 \\ 0 & 0 & 1 \end{bmatrix} \quad (9)$$

$$R = R_z(\beta)R_y(\alpha)R_x(\theta) \qquad (10)$$

Assuming $(x, y, z, 1)$ and $(x^t, y^t, z^t, 1)$ are the homogeneous coordinates of point $Q$ and $P$, $(x', y', z', 1)$ is the homogeneous coordinate of $Q$ after the rotation transformation, the rotation transformation follows:

$$\begin{bmatrix} x' \\ y' \\ z' \\ 1 \end{bmatrix} = \begin{bmatrix} R & 0 \\ 0 & 1 \end{bmatrix} \begin{bmatrix} x \\ y \\ z \\ 1 \end{bmatrix} \qquad (11)$$

Assuming that the $(t_x, t_y, t_z, 1)$ is a translation vector, we can obtain the function after rotation and translation transformation as follow:

$$\begin{bmatrix} x^t \\ y^t \\ z^t \\ 1 \end{bmatrix} = \begin{bmatrix} 1 & 0 & 0 & t_x \\ 0 & 1 & 0 & t_y \\ 0 & 0 & 1 & t_z \\ 0 & 0 & 0 & 1 \end{bmatrix} \begin{bmatrix} R & 0 \\ 0 & 1 \end{bmatrix} \begin{bmatrix} x \\ y \\ z \\ 1 \end{bmatrix} = M \begin{bmatrix} x \\ y \\ z \\ 1 \end{bmatrix} \qquad (12)$$

## 3. Urban data acquisition system combining UAV and mobile mapping vehicle

In this study, we proposed an urban data acquisition system combining the observations from UAV and mobile mapping vehicle. UAV is responsible for collecting aerial images, while mobile mapping vehicle is responsible for collecting ground-view data. Their combination achieves fast acquisition of three-dimensional spatio-temporal information in complex urban environments.

### 3.1. UAV sub-system

Traditional platforms to obtain images are manned aircraft or satellites (Woldai 2020). However, these remote sensing systems have several major disadvantages, such as low spatial and temporal resolutions, limited availability by weather conditions, and high costs (Xiang and Tian 2011). In comparison, UAVs are typically at a low cost. Such light-weighted aircraft with low speed has shown great performance in remote sensing data collection. UAV-based sensing system largely fills the gap between ground observations and remote observations from satellite platforms. UAV Remote Sensing System (UAVRSS) is a new remote sensing system composed of an unmanned aerial platform, Positioning and Orientation System (POS), remote sensing sensor, and inertial stabilization platform. It can automatically, intelligently, and rapidly obtain imagery that covers targeted areas and can process, model, and analyze the obtained images (Li, Shan, and Gong 2009). The overall framework of the UAV system is shown in Figure 2. The POS is used to provide the position and attitude information and assist the high-resolution imaging. The main types of sensors used for UAV remote sensing include aerial cameras, airborne LiDAR (Horvat et al. 2016), hyperspectral imagers (Zhong et al. 2018), infrared thermal imager, and small Synthetic Aperture Radar (SAR) (Guerreiro et al. 2017). In the case of UAV flight, the inertial stabilization platform is applied to ensure the stability of remote sensing sensors by absorbing and smoothing the mechanical jitter during flight.

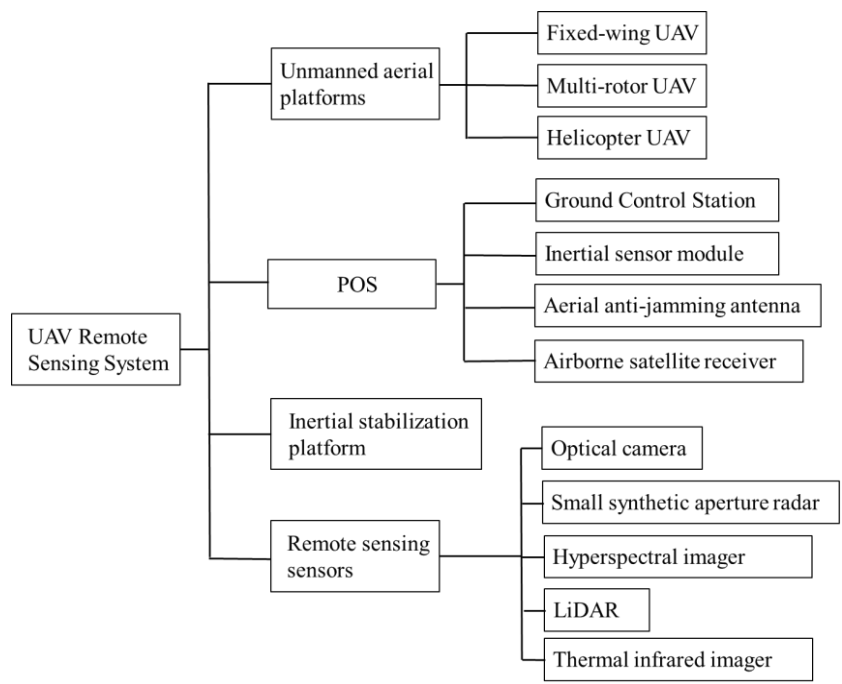

Figure 2. The overall framework of the UAV Remote Sensing System.

1) **Unmanned aerial platforms**

UAV is the flight platform of photogrammetry and remote sensing system, which is used to carry remote sensing sensors, positioning and attitude measurement system, UAV power system, and other remote sensing measurement related equipment. The most important function of the flight platform is to carry a variety of sensors to carry out safe and stable flight tasks in order to ensure the acquisition of high-quality remote sensing data. Innovations in power technology, lightweight composite materials, and control methods have led to the emergence of a variety of UAVs.

UAVs can be generally divided into three categories based on their structure: 1) fixed-wing UAVs, 2) multi-rotor UAVs, and 3) helicopter UAVs, as shown in Figure 3. Fixed-wing UAVs are mainly used in military and civil fields with the advantages of long endurance time and strong load capacity and the disadvantages of strict requirements for taking off and landing. The multi-rotor UAV, mainly used in the civil field, is able to rise and fall vertically and hover in the air. Helicopter UAVs can also achieve vertical lifting and hovering in the air with high load capacity compared to multi-rotor UAVs.

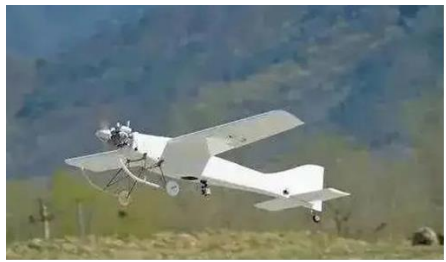 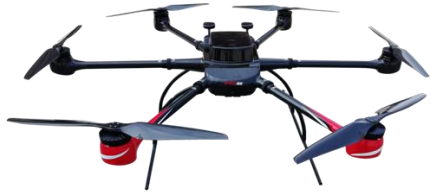 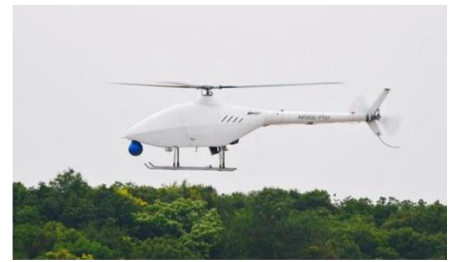

(a) Fixed-wing UAV  (b) Multi-rotor UAV  (c) Helicopter UAV

Figure 3. Different types of UAVs.

2) **POS**

POS, used to calculate the position and attitude parameters of remote sensing sensor, is composed of a Global Navigation Satellite System (GNSS) and Inertial Navigation System (INS). POS mainly includes POS integration module, aerial anti-jamming antenna, ground control station, and POS post-processing software. POS receives satellite signals as well as angular velocity and acceleration collected by the IMU sensor with the support of GNSS antenna the ground control stations. POS post-processing software is used for fusion calculation, aiming to obtain the three-dimensional coordinates and attitude of the carrier.

3) **Remote sensing sensors**

Given different tasks, the remote sensing sensor uses the corresponding airborne remote sensing equipment, such as high-resolution CCD digital camera, light optical camera, multispectral imager, infrared scanner, laser scanner, hyperspectral imager, synthetic aperture radar, etc. (Colomina and Molina 2014). Remote sensing sensors should own the characteristics of small size, light weight, high precision, and large storage capacity.

The aerial camera shoots high-resolution optical images for aerial remote sensing. The hyperspectral imager combines imaging technology and spectral technology to obtain continuous and narrow reflectance of the spectrum. According to the laser ranging principle, laser scanners acquire three-dimensional coordinates and texture information of a large number of dense points on the surface of objects and construct the three-dimensional model of the object and various map data such as line, plane, and volume. Infrared scanners sense the infrared radiation of the measured object and form an infrared image by combining the optical scanning and motion direction of the instrument.

4) **Inertial stabilization platform**

In general, the volume of UAVs is relatively small. Therefore, during flight, UAVs are vulnerable to interferences caused by weather conditions such as crosswind and eddy current, potentially leading to error measurement of IMU and resulting in decreased imaging quality. The inertial stabilization platform supports and stabilizes the navigation and positioning sensors and remote sensing sensors, which can effectively isolate the angular movement of the flight platform and the errors caused by various internal and external disturbances and maintains the working stability of the POS and remote sensing sensors.

In this study, we use the light, small, and high-precision POS UAV remote sensing system independently developed by Leador Spatial Information Technology Corporation (Figure 4). Its unmanned aerial platform is a professional six-rotor UAV with excellent dynamic redundancy and wind resistance. It can be adapted to work in high-altitude areas and has a considerably long endurance time of more than 50 min, which is suitable for long-term operation. With the support of ground control stations, the UAV can be set in automatic cruise, automatic flight, automatic landing, intelligent cruise. Thanks to its highly modulated design and pluggable structure, the entire UAV can be stored in an aeronautical box for easy transportation.

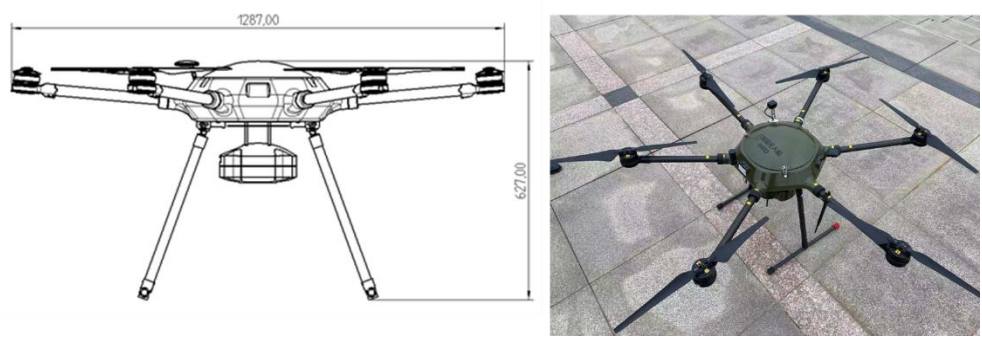

Figure 4. The professional six-rotor UAV developed by Leador Spatial Information Technology Corporation.

*3.2 Mobile mapping vehicle sub-system*

The mobile mapping vehicle system is one of the most cutting-edge science and technology of modern surveying and mapping, integrated with the global satellite positioning, inertial navigation, image processing, photogrammetry, laser scanning, geographic information, and integrated control technology, with the characteristics of flexibility, high precision, high resolution, and real-time multi-source 3D spatial data collection. Mobile mapping vehicles can obtain essential surface information of roads as well as the vertical objects (e.g., buildings and trees) on both sides of the road in a real-time manner, even under fast driving conditions. This surface information reflects the structure, size, texture, and other information of urban objects. Mobile mapping vehicle systems have obvious advantages in data acquisition: fast, accurate, automatic in data processing workflow with the adaptation of various data forms.

The architecture of the acquisition, processing, and application of the mobile mapping vehicle system can be divided into four levels: equipment layer, data layer, outcome layer, and application layer (Figure 5).

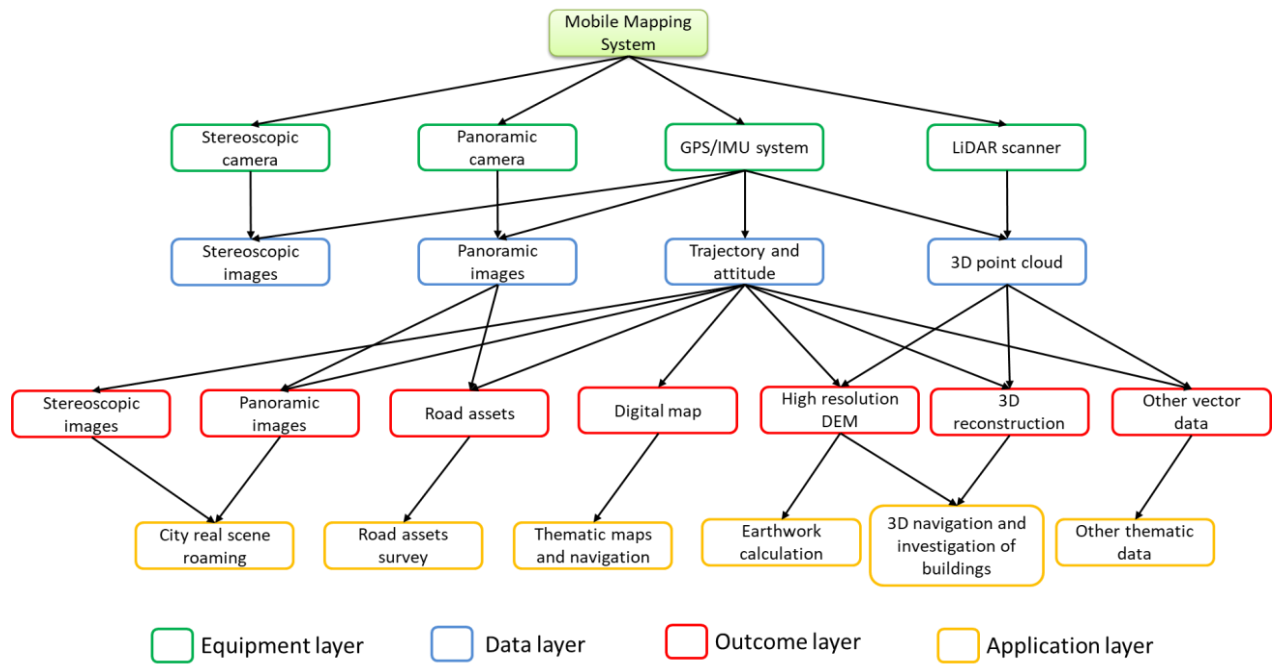

Figure 5. The architecture of the acquisition, processing, and application of a mobile mapping vehicle system.

The equipment layer contains on-board data acquisition hardware equipment, including POS that serves as the time and space reference, panoramic cameras used to obtain image data, and laser scanners that can directly obtain 3D information.

The data layer contains data collected directly by the device or obtained after simple processing. Among them, POS can recover high-precision running track and position and attitude with high sampling frequency. Positioning and attitude information can be used to mark the position and azimuth of panoramic images, serving as time and space references for 3D point clouds.

The outcome layer represents the information of interest obtained from the data layer. For example, high-precision driving tracks can be used as digital maps to update the road network. The point cloud contains abundant information regarding the physical size of the object, which can be used to construct high-resolution DEM of the road surface and roadside, three-dimensional building model, and positions of power lines, poles, and other ancillary facilities.

The application layer serves as the application of the outcome layer. Different industries have different demands with specific applications. For example, panoramic images can be used for real virtual roaming and navigation, to identify the position of advertisements on the street, and to calculate earthwork volume for road diversion based on high-resolution DEM.

The mobile mapping vehicle system used in this study, shown in Figure 6, is the "Flash" system with an exquisite modular design developed by Leador Spatial Information Technology Corporation. Connected through aviation plugs, the whole system is composed of a data acquisition device, monitoring device, and power supply. The data acquisition device is composed of high-precision optical fiber or laser Inertial Navigation System (INS), panoramic camera, LiDAR, and high-grade protective cover. The INS, panoramic camera, and LiDAR are fixated with mechanical devices. Indoor integrated checking and calibration are completed before delivery. The monitoring device is supported by an industrial computer, which is used for the system's working state display and monitoring, data storage, processing, etc.

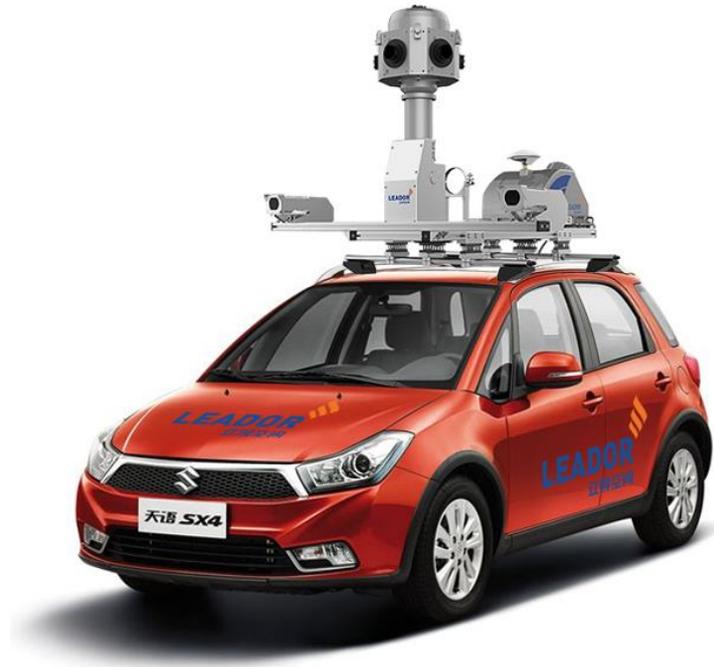

Figure 6. The mobile mapping vehicle system developed by Leador Spatial Information Technology Corporation.

*3.3 The combination of UAV and mobile mapping vehicle system*

Figure 7 presents how UAV and mobile mapping vehicle systems are integrated. UAV and mobile mapping vehicle can be equipped with different sensors, depending on project requirements. The aerial images collected by UAV and the ground images collected by mobile mapping vehicle own the characteristics of multiple spatial resolution, temporal resolution, spectral resolution, and multi angles. The proposed spatial-temporal-spectral-angular observation model is used for data fusion to achieve rapid production of high-precision and high-quality 3D products in complicated urban environments.

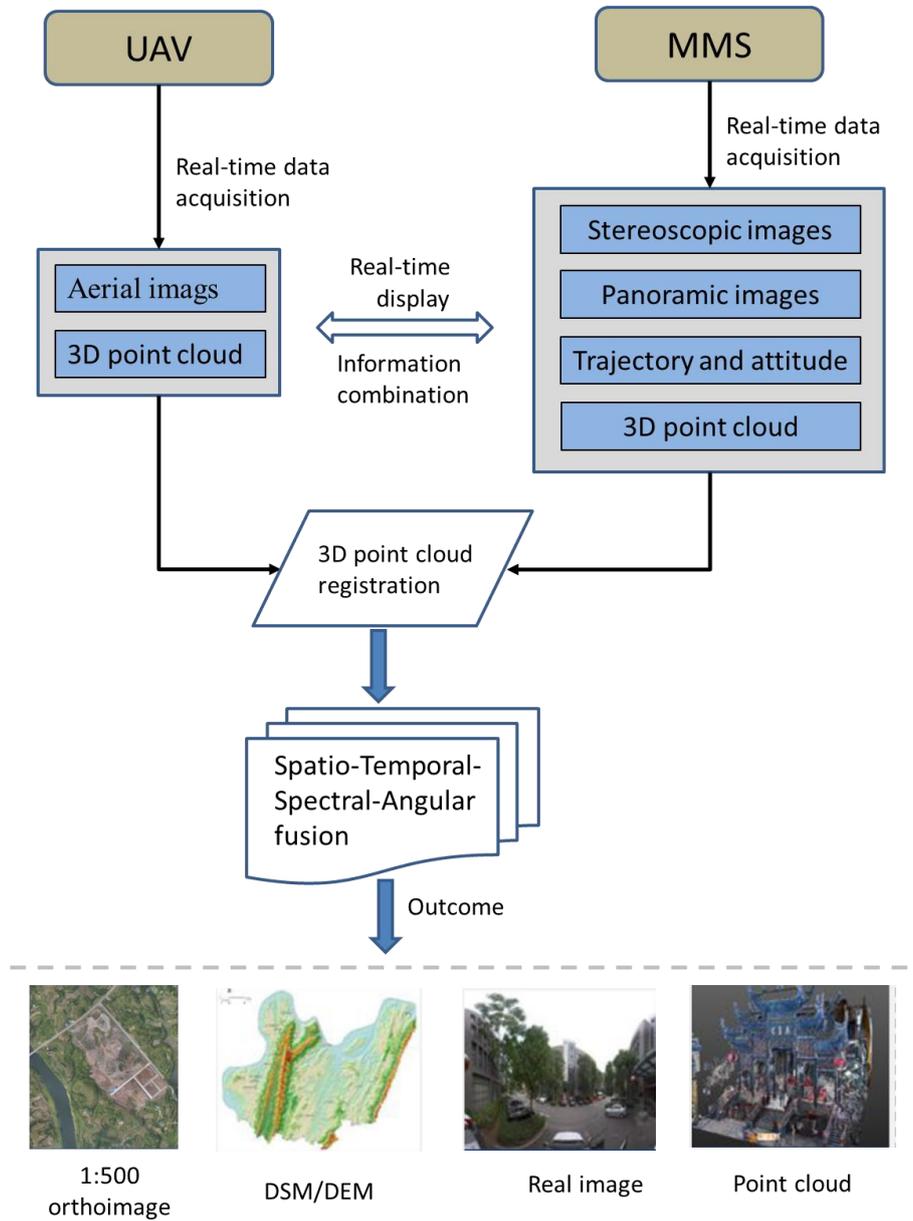

Figure 7. The combination of UAV and mobile mapping vehicle system.

In this study, the UAV is equipped with five Sony cameras with three bands (Red, Green, Blue), obtaining images from diversified angles. The mobile mapping vehicle captures the ground data, including panoramic images, 3D point clouds, and trajectory data. We conduct the voxel-level fusion that combines the 3D point cloud data from UAV and mobile mapping vehicle. The UAV 3D point cloud data is first extracted from UAV images with accurate pose and position information. We then perform the registration of 3D point cloud data from UAV and mobile mapping vehicle, including three steps: 1) keypoints selection, 2) transformation matrix calculation, and 3) aligning. Finally, we implement the proposed spatio-temporal-spectral-angular observation model.

## 4. Experiments

### 4.1 Study area

The study area is in Baisha Town, Chongqing, China, located at 106°7'1.55" E and 29°3'38.66" N. Baisha Town is featured by complicated urban environments with occlusions caused by trees and buildings where a single sensor often fail to obtain complete observations. Therefore, we conduct the spatio-temporal-spectral-angular observation based on the combination of UAV and mobile mapping vehicle system in this study area, as shown in Figure 8. One UAV and one mobile mapping vehicle system are utilized in our experiments. The red line indicates the data acquisition area where the mobile mapping vehicle obtained data along the road and the UAV collected data from the air.

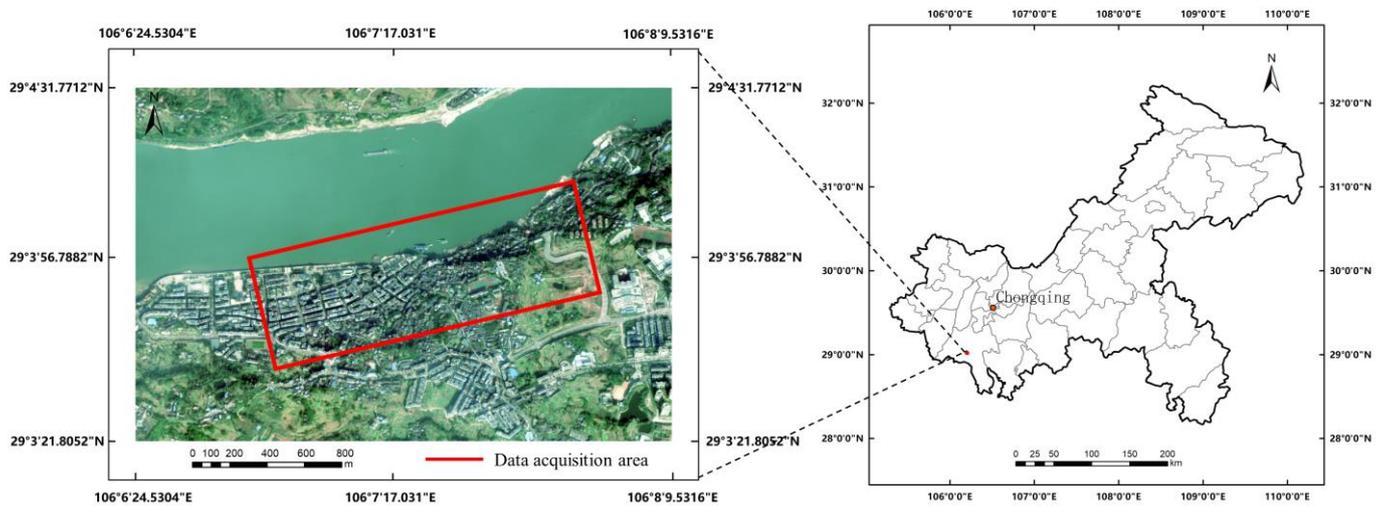

Figure 8. The experimental area is in Baisha Town, Chongqing, China, with the red line indicating the data acquisition area.

### 4.2 Data acquisition

In the study area, the UAV and mobile mapping vehicle collected data simultaneously according to the prescribed routes, as shown in Figure 9. The UAV captured images from diversified angles while the mobile mapping vehicle obtained images and 3D point cloud data along the road.

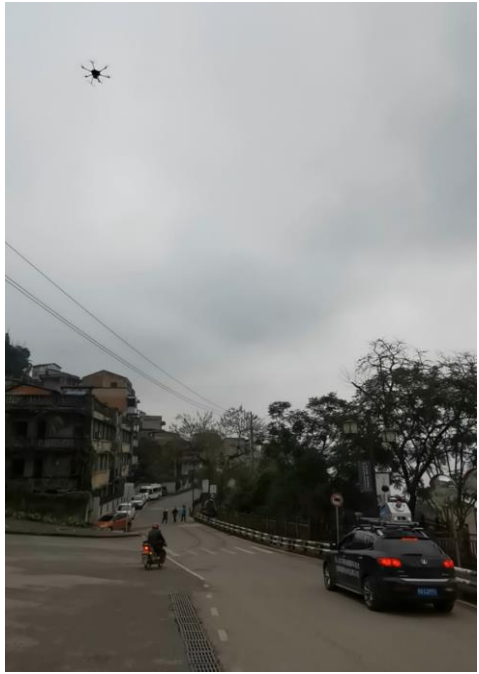

Figure 9. The UAV and mobile mapping vehicle were collecting data in a simultaneous manner with prescribed routes.

The UAV, equipped with five Sony cameras (SONY ILCE-5100), can capture the ground images from five angles. The average altitude of the flight was set to 50 m. The UAV obtained a total of 135 images, with 27 images per camera. The size of each image is 6000×4000. Figure 10 shows some example images from our UAV. The positional information of each image was recorded, an example showing in Table 1.

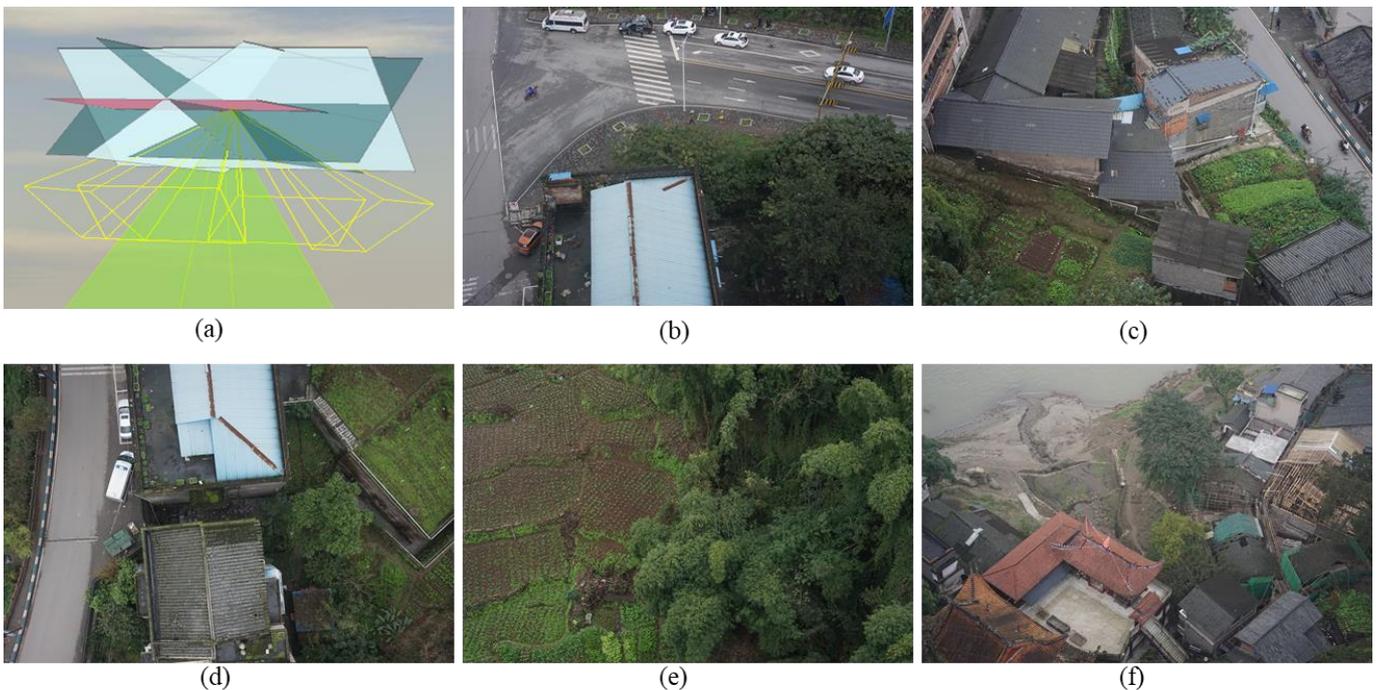

Figure 10. The UAV images from five cameras. (a) Five-lens camera model, (b) The behind view, (c) The forward view, (d) The vertical view, (e) The left side view, (f) The right side view.

Table 1. The position information of part of the sample from UAV.

| Id | Latitude (°) | Longitude (°) | Altitude (m) | Yaw(°) |
|---|---|---|---|---|
| 1 | 29.06586 | 106.1266 | 50.08 | 259.1 |
| 2 | 29.06585 | 106.1265 | 49.99 | 249.5 |
| 3 | 29.06582 | 106.1264 | 50.1 | 248.6 |
| 4 | 29.0658 | 106.1264 | 49.96 | 248.6 |
| 5 | 29.06578 | 106.1263 | 49.97 | 248.7 |
| 6 | 29.06577 | 106.1262 | 50.01 | 248.5 |
| 7 | 29.06576 | 106.1262 | 50.02 | 248.7 |
| 8 | 29.06575 | 106.1261 | 49.95 | 248.8 |
| 9 | 29.06572 | 106.126 | 49.96 | 249.4 |
| 10 | 29.06569 | 106.126 | 49.94 | 249.7 |

The mobile mapping vehicle moved with a speed of 30 km/h and captured the ground data that include panoramic images, 3D point cloud, trajectory data. Figure 11 shows an example of data collected from our mobile mapping vehicle.

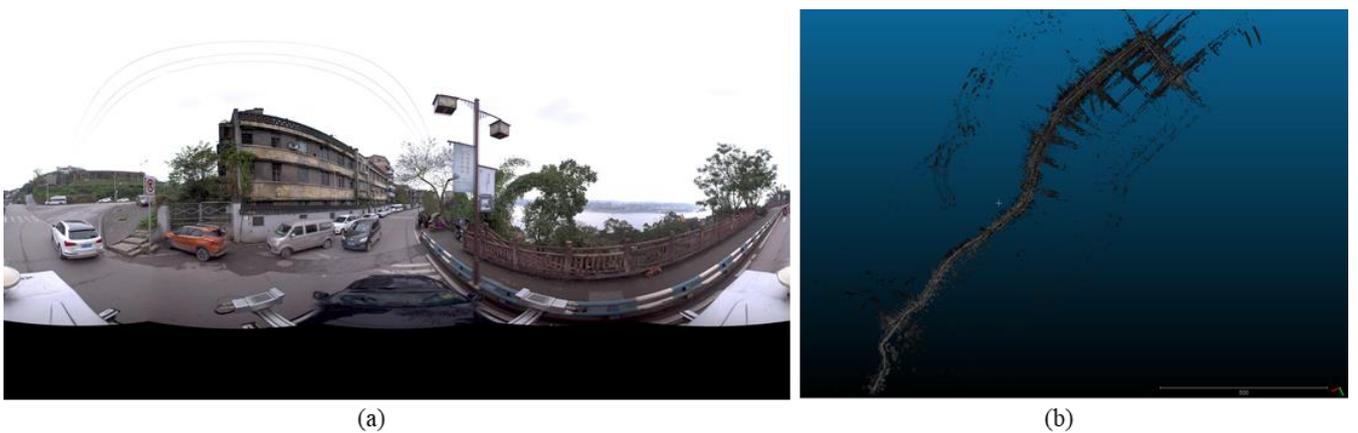

(a)     (b)

Figure 11. Data sample from our mobile mapping vehicle. (a) Panoramic image, (b) 3D point cloud.

*4. 3 Data preprocessing and data fusion*

In the data preprocessing step, we first extracted the 3D point cloud from UAV images according to the accurate pose and position information of each image. Figure 12 shows the extraction results. UAV 3D point cloud has rich color information and spatial texture features of ground objects, as well as spatial location information, contributing to ground object identification and 3D model information extraction. Although rich information can be obtained from the UAV point cloud, there still remains missing information, such as information from facades and occluded objects.

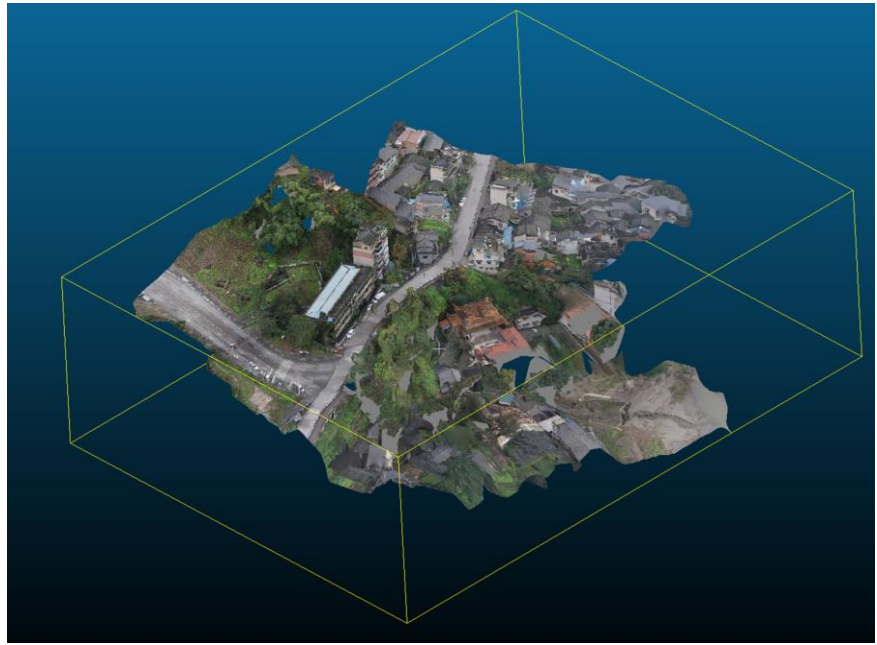

Figure 12. 3D point cloud extracted from UAV images.

To combine the 3D point clouds from UAV and mobile mapping vehicle, point cloud registration was conducted to transform these two kinds of point clouds to the same coordinate system. This operation was based on Cloud Compare software. Considering the enormous size of the 3D point cloud data, we utilized data from only a small area for experiments. To align these two data sources, we set the MMS 3D point cloud as the reference and aligned the UAV 3D point cloud to the reference. We chose seven pairs of keypoints between these two data sources, as shown in Figure 13, and then calculated the rigid transformation matrix (*M*) (Equation.13). The final Root Mean Square Error (RMSE) was 0.6001.

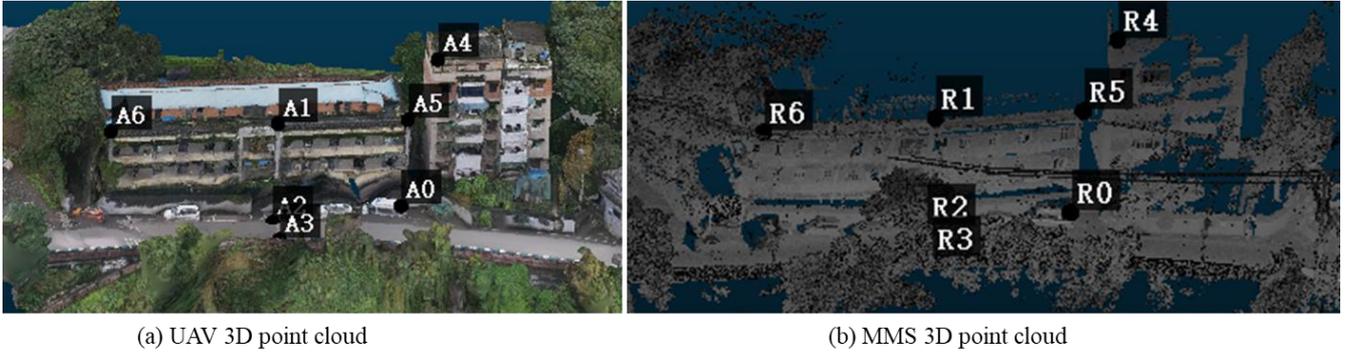

(a) UAV 3D point cloud  (b) MMS 3D point cloud

Figure 13. Keypoints selection between the UAV 3D point cloud and the MMS 3D point cloud.

$$\begin{bmatrix} 0.953 & -0.016 & -0.019 & -286.686 \\ 0.016 & 0.953 & 0.013 & -85.317 \\ 0.019 & -0.013 & 0.953 & -3.685 \\ 0.000 & 0.000 & 0.000 & 1.000 \end{bmatrix} \quad (13)$$

Finally, we integrated the aligned two data sources based on our proposed spatio-temporal-spectral-angular observation model to generate a new integrated 3D point cloud that combines information from UAV and MMS point clouds (Figure 14). The UAV point cloud contains rich color information and spatial texture features but lacks facade information, which can be supplemented by the MMS point cloud. The integrated 3D point cloud achieves urban observation in a comprehensive manner.

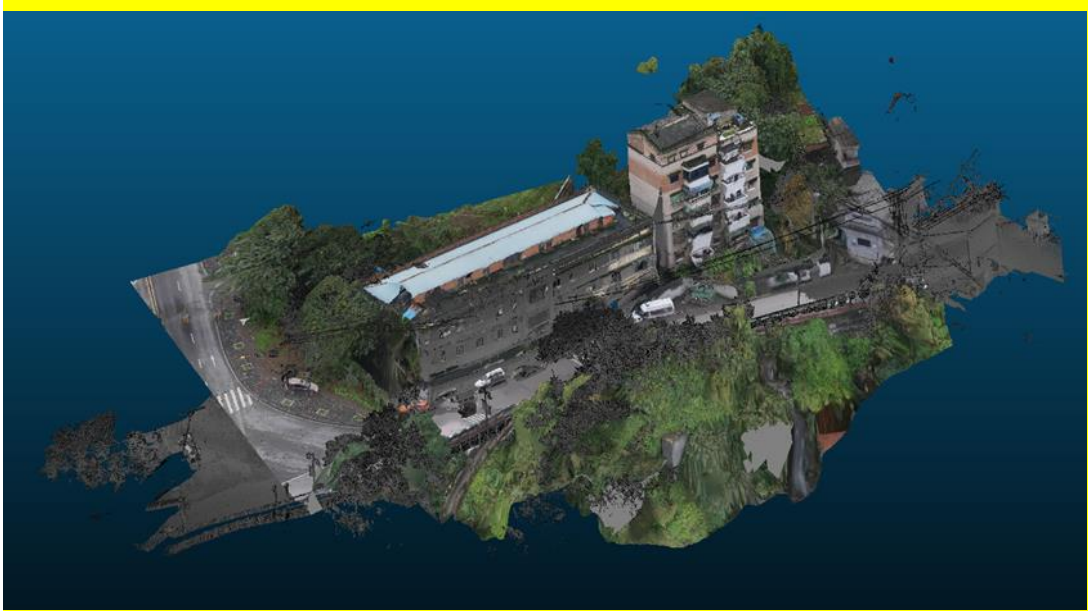

Figure 14. The integrated 3D point cloud that combines the UAV point cloud and the MMS point cloud.

*4. 4 Analysis*

Figure 15 (a) and (b) present the top views of UAV and MMS point cloud. We can observe that the UAV point cloud can capture more abundant top surface information of ground objects with rich color information and texture features. In comparison, the MMS point cloud fails to obtain top surface information of buildings (red box) as well as information about the occluded parts of ground objects (yellow box). Figure 15 (c) and (d) present the side views of UAV and MMS point cloud. It can be seen that the UAV point cloud contains notable distortion with blurry (green box) and missing information (red box) and fails to accurately obtain the facade information of ground objects. Figure 15 (e) and (f) present the integrated 3D point cloud that combines the UAV point cloud and the MMS point cloud. The fused data combines the advantages of UAV and mobile mapping vehicle, containing rich top surface information (color and texture) from UAV and rich facade information (texture details) from mobile mapping vehicle, thus solving the problem of data missing caused by occlusion. Therefore, from the comparison results, we can conclude that information from a single sensor is inadequate to obtain the 3D spatial information of ground objects in complex urban scenes in an accurate manner. The integrated point cloud data contains more abundant and comprehensive information, suggesting that the data fusion strategy in our spatio-temporal-spectral-angular model is able to obtain 3D information for complex urban environments in a comprehensive manner.

## 5. Discussion

As a conceptual model, the proposed spatio-temporal-spectral-angular observation model includes pixel-level (or voxel-level) fusion and feature-level fusion that enables the integration of similar data types. In this study, we conducted voxel-level fusion that aims to integrate UAV and MMS 3D point clouds. We acknowledge that certain improvements can be made in future works. In our experiments, the UAV 3D point cloud was extracted from UAV images instead of directly from UAV equipped with a LiDAR sensor. Such a point cloud extraction method leads to a relatively lower precision. In addition, the MMS point cloud we collected lacks color information. In light of these two limitations, we only conducted a qualitative evaluation of our data fusion strategy instead of a quantitative one. Despite these limitations, the spatio-temporal-spectral-angular observation model we proposed achieves the integration of multi-sensor, multi-angle data to better obtain three-dimensional information of complex urban environments. Our experiment results confirmed the effectiveness and contribution of our spatio-temporal-spectral-angular observation model. In the future, we plan to further investigate the performance of the proposed spatio-temporal-spectral-angular observation model, applying it to other scenarios and paying more attention to feature-level fusion and experiments based on UAV and MMS equipped with better LiDAR sensors.

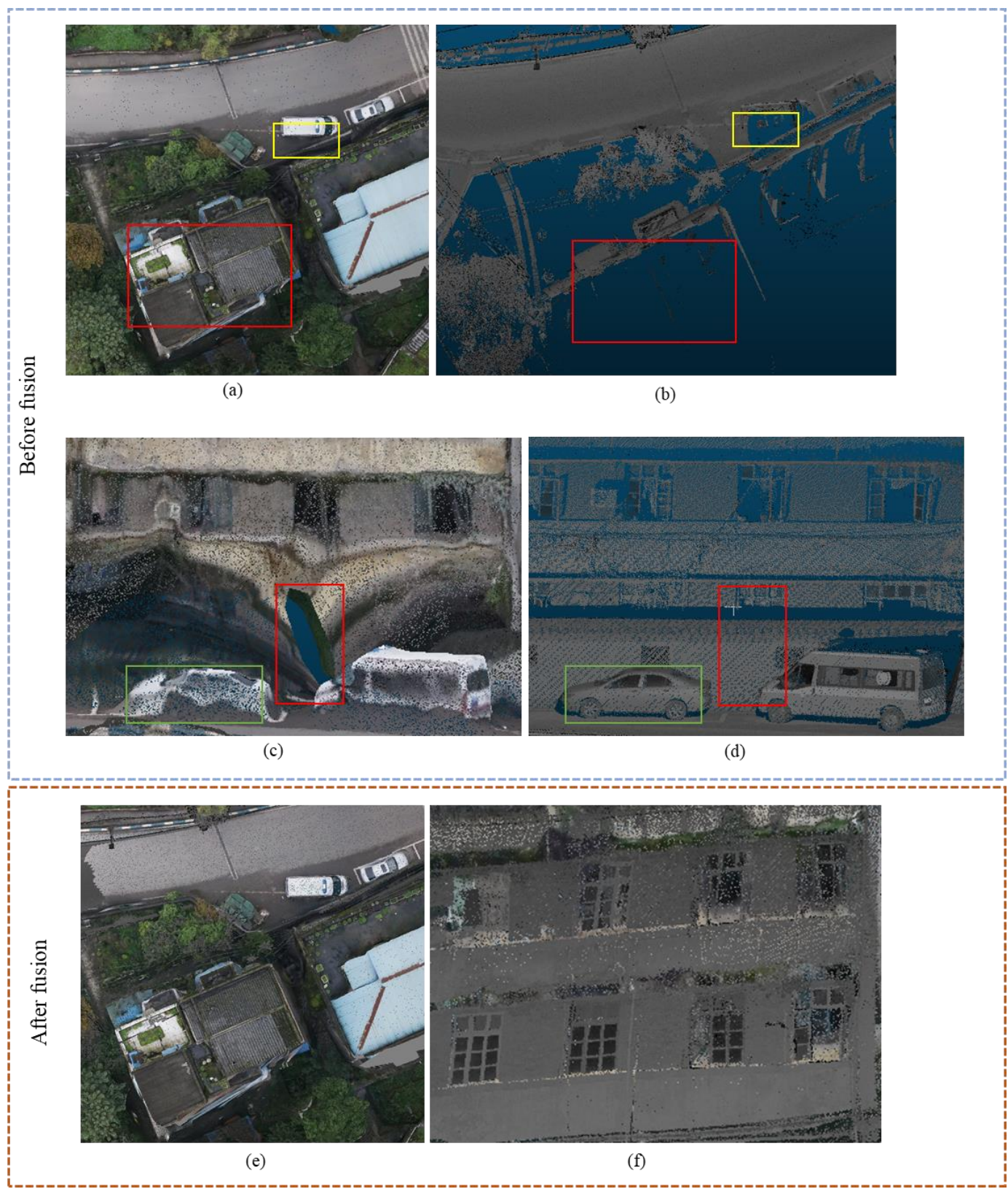

Figure 15. The comparison results of 3D point cloud before and after fusion. (a) and (c) present the UAV point cloud. (b) and (d) present the MMS point cloud. (e) and (f) present the integrated 3D point cloud by fusing UAV and MMS point

clouds.

## 6. Conclusions

In this study, we propose a spatio-temporal-spectral-angular observation model to integrate observations from UAV and mobile mapping vehicle platforms, realizing a joint, coordinated observation operation. We develop a multi-source remote sensing data acquisition system to effectively acquire multi-angle data of complex urban scenes. Multi-source data fusion solves the missing data problem caused by occlusion and achieves accurate, rapid, and complete collection of holographic spatial and temporal information in complex urban scenes. We carried out an experiment on Baisha Town, Chongqing, China and obtained multi-sensor, multi-angle data from UAV and mobile mapping vehicle system. We first extracted the point cloud from the UAV and then integrated the UAV and mobile mapping vehicle 3D point clouds. The experimental results show that a single sensor is unable to accurately obtain the 3D spatial information of ground objects while the combination of information from multiple sensors can address this issue. The integrated point cloud contains more abundant and comprehensive information, indicating that the data fusion via our proposed spatio-temporal-spectral-angular model can better obtain the 3D information for complex urban environments, providing an effective data acquisition solution towards comprehensive urban monitoring.

**Funding**

This work is supported by the National Key Research and Development Program of China [grant number 2018YFB2100501], the National Natural Science Foundation of China [grant numbers 42090012, 41771452, 41771454, and 41901340].

**Notes on contributors**

*Zhenfeng Shao* received the PhD degree in photogrammetry and remote sensing from Wuhan University in 2004. Since 2009, he has been a Full Professor with the State Key Laboratory of Information Engineering in Surveying, Mapping and Remote Sensing (LIESMARS), Wuhan University. He has authored or coauthored over 70 peer-reviewed articles in international journals. His research interests include high-resolution image processing, pattern recognition, and urban remote sensing applications. Since 2019, he has served as an Associate Editor of the Photogrammetric Engineering &

**Acknowledgements**

The authors are sincerely grateful to the editors as well as the anonymous reviewers for their valuable suggestions and comments that helped us improve this paper significantly. Thanks sincerely to Jiaming Wang, Zhiqiang Wang, Linze Bai, Sihang Zhang, and Lan Ye for their help with this paper.

## Data Availability Statement

The data that support the findings of this study are available from the corresponding author, upon reasonable request.